\documentclass[letterpaper, 10 pt, conference]{ieeeconf}
\IEEEoverridecommandlockouts
\IEEEtriggeratref{12} 

\usepackage{xcolor}
\usepackage{amssymb} 
\usepackage{amsmath}
\usepackage{subcaption}
\usepackage{float}
\usepackage{booktabs}
\usepackage{graphicx}
\usepackage{tikz} 
\usetikzlibrary{arrows,decorations.markings}
\usepackage{cite}
\usepackage{authblk}
\usepackage{gensymb}
\usepackage{array}
\newcolumntype{P}[1]{>{\centering\arraybackslash}p{#1}}
\usepackage{flushend}
\usepackage{hyperref}

\title{\bf DFBVS: Deep Feature-Based Visual Servo}

\author{Nicholas Adrian, Van-Thach Do, and Quang-Cuong Pham%
  \thanks{The authors are with the HP-NTU Digital Manufacturing
  Corporate Lab and School of Mechanical and Aerospace Engineering,
  Nanyang Technological University, Singapore. Email:
  nicholasadr@ntu.edu.sg}}

\begin{document}

\maketitle
\thispagestyle{empty}
\pagestyle{empty}

\begin{abstract}
  Classical Visual Servoing (VS) rely on handcrafted visual features,
  which limit their generalizability. Recently, a number of
  approaches, some based on Deep Neural Networks, have been proposed
  to overcome this limitation by comparing directly the entire target
  and current camera images. However, by getting rid of the visual
  features altogether, those approaches require the target and current
  images to be essentially similar, which precludes the generalization
  to unknown, cluttered, scenes. Here we propose to perform VS based
  on visual features as in classical VS approaches but, contrary to
  the latter, we leverage recent breakthroughs in Deep Learning to
  automatically extract and match the visual features. By doing so,
  our approach enjoys the advantages from both worlds: (i) because our
  approach is based on visual features, it is able to steer the robot
  towards the object of interest even in presence of significant
  distraction in the background; (ii) because the features are
  automatically extracted and matched, our approach can easily and
  automatically generalize to unseen objects and scenes. In addition,
  we propose to use a render engine to synthesize the target image,
  which offers a further level of generalization. We demonstrate these
  advantages in a robotic grasping task, where the robot is able to
  steer, with high accuracy, towards the object to grasp, based simply
  on an image of the object rendered from the camera view
  corresponding to the desired robot grasping pose.
\end{abstract}

\section{Introduction}
\label{part:introduction}

Visual Servoing (VS), where images recorded from a camera mounted on
the robot end-effector are used to iteratively guide the robot
motions, is a key technique to tackle robotic tasks requiring high
accuracy~(see e.g. \cite{chaumette2006visual} for a review). Classical
approaches to VS require handcrafting visual features, such as points,
lines, or corners, which limit their generalizability.

Recently, a number of approaches have been proposed to overcome the
reliance of classical VS on handcrafted
features. In~\cite{collewet2008visual,collewet2011photometric},
the authors considered the whole image as a single feature
to circumvent the need to construct and extract special features. In~\cite{saxena2017exploring,bateux2018training,yu2019siamese,tokuda2021convolutional},
Convolutional Neural Networks (CNNs) are used on whole images, either
to directly estimate the pose difference between the target and
current image, or to find the instantaneous robot control in an
end-to-end manner.

\begin{figure}[!t]
  \centering
  \includegraphics[width=\columnwidth]{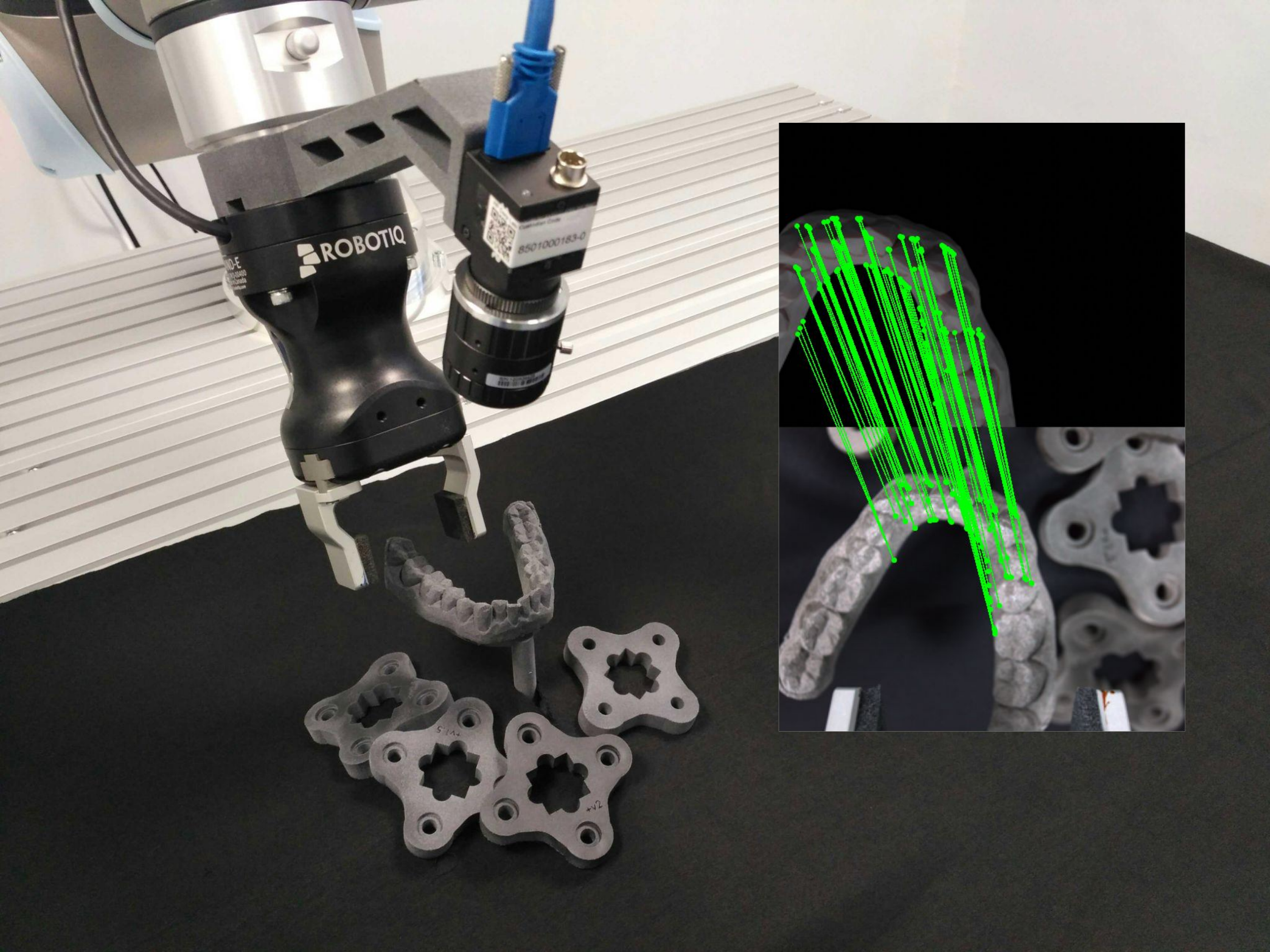}
  \caption{Visual Servoing on a 3D printed object. A Deep Neural
    Network (\emph{not} trained on the object at hand)
    outputs rich feature points separately on two images: target (top)
    and current (bottom) images, forming feature correspondences
    despite dissimilar backgrounds. The target image is rendered from
    the object's CAD model viewed from the desired grasping pose.
    The video of the experiments is available at \url{https://youtu.be/vFAdebgog7k}.}
  \label{fig:visual_id}
\end{figure}

We argue that getting rid of the visual features altogether is akin to
throwing the baby out with the bathwater. Indeed, by comparing
directly the entire target and current images, the approaches just
mentioned require those two images to be essentially similar, which
precludes the \emph{generalization to unknown, cluttered,
  scenes}. Consider for instance the situation of
Fig.~\ref{fig:visual_id}, where the object of interest is isolated in
the target image, but surrounded by distractors in the current
image. Approaches based on ``whole-image'' processing will
fundamentally be unable to steer the robot towards the target camera
view of the object of interest because the two images are too
different.

On the contrary, a feature-based approach will enjoy an
\emph{attention-like} mechanism: if one can match a sufficient number
of features \emph{on the object of interest} in the target and current
images, Visual Servoing will be possible even in the presence of
significant distraction. Building on this insight, we propose here to
perform VS based on visual features as in classical VS approaches. But
contrary to the latter, we leverage recent breakthroughs in Deep
Learning~\cite{ono2018lf,detone2018superpoint,christiansen2019unsuperpoint,tang2020neural}
to \emph{automatically} extract and match those visual features
between the current and the target images. Our approach therefore
enjoys the following advantages as compared to the literature:
\begin{itemize}
\item because our approach is based on visual features, it is able to
  steer the robot towards the object of interest even in presence of
  significant distraction in the background;
\item because the features are automatically extracted and matched,
  our approach can easily and automatically generalize to unseen
  objects and scenes.
\end{itemize}

\begin{figure*}[!t]
  \centering
  \vspace{1mm}
  \includegraphics[width=\textwidth]{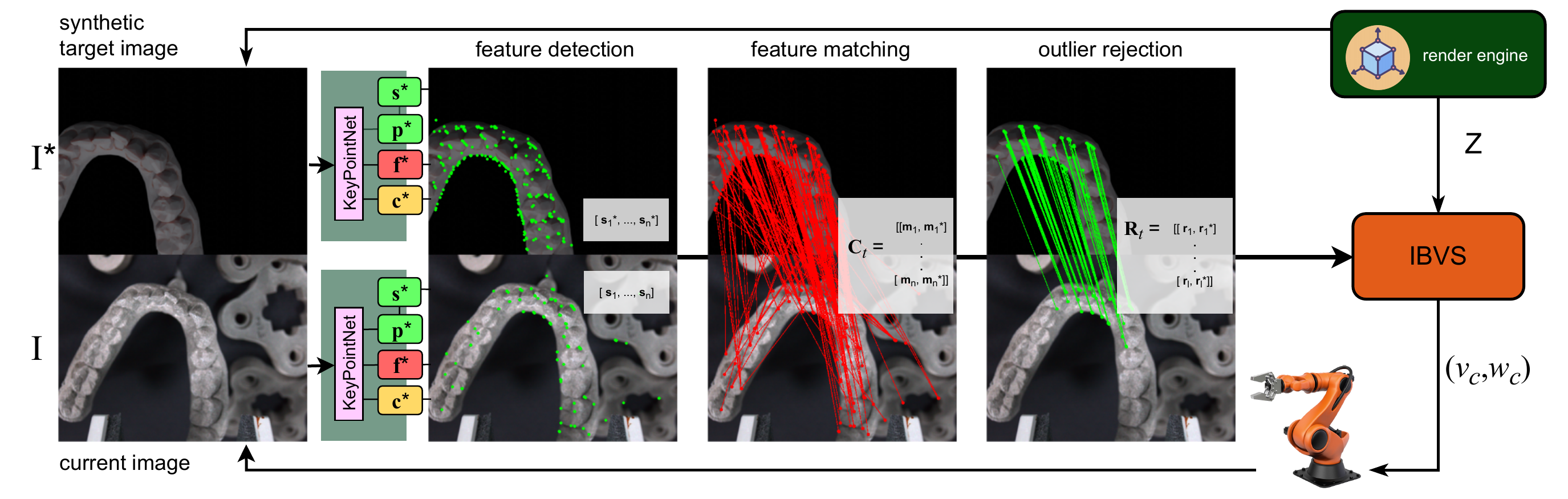}
  \caption{Our proposed pipeline for visual servoing with DL-based
    sparse feature detection (Section \ref{part:dfbvs-detect}),
    feature matching (Section \ref{part:dfbvs-match}), and outlier
    rejction (Section \ref{part:dfbvs-match}). Render engine (Section
    \ref{part:dfbvs-render}) renders the target image and
    provides accurate depth values as part of the required input to
    IBVS control law (Section \ref{part:vs-intro}).}
  \label{fig:flowchart}
\end{figure*}

In addition, we propose to use a render engine to synthesize the
target image. This enables a simple and automatic way to acquire the
target image, while providing, moreover, accurate depth values for the
visual features. Note that this ``sim-to-real'' idea is made possible
only by the attention-like mechanism discussed previously. Figure
\ref{fig:flowchart} shows the flowchart of the overall pipeline.

The remainder of the paper is organized as follows. In Section
\ref{part:lit-review}, we review related works in visual servoing,
emphasizing on methods that utilize Deep Neural Network (DNN). In
Section \ref{part:vs-intro}, we present brief theoretical background
behind Image-Based Visual Servo (IBVS) control which our method relies on.
The components of DFBVS are elaborated in Section
\ref{part:dfbvs}. In Section \ref{part:exp} we show that our
method can achieve accurate positioning (0.89 pixels average error)
and convergence on three unseen objects despite significant
distraction on the current image. Finally, we conclude in
Section \ref{part:conclusion}.

\section{Literature Review}
\label{part:lit-review}

\subsection{Classical Visual Servo}

In general, classical visual servoing consists of two classes:
Image-Based Visual Servo (IBVS) and Position-Based Visual Servo (PBVS)
\cite{chaumette2006visual}. In
IBVS~\cite{feddema1989vision,weiss1987dynamic}, the features are
directly obtained from the image measurement. While in PBVS
\cite{wilson1996relative}, the image measurement is further developed
into 3D parameters. Generally, PBVS is more computationally demanding
due to the need to establish 2D-3D correspondence for each feature and
thus requires camera intrinsic parameters and the object's 3D model
while IBVS only requires the former (this simplicity in feature calculation
is the reason why we base our method on IBVS).
PBVS control scheme theoretically
can achieve global asymptotic stability but under the strict
requirement for correct calculation of 3D parameters, else it affect
the final accuracy. On the other hand, IBVS has local asymptotic
stability at best but is more robust towards errors in calibration and
depth estimation. In both classes, classical visual servoing methods
rely on extracting hand-crafted visual features, which is error-prone
and has limited generalizability.

Direct visual servoing (DVS)~\cite{collewet2008visual,collewet2011photometric}
was designed to circumvent the need for image feature extraction.
Instead, analytical calculation of interaction matrix based on
illumination model is used to link temporal variation of whole
image luminance values with image motion. While it can achieve
lower positioning error compared to classical visual servoing,
it has smaller convergence domain. 

\subsection{Deep Learning (DL) for Visual Servo}

DL is increasingly explored to improve existing visual servoing methods.
Most approaches wager that such improvement can be obtained through
dropping reliance on extracting visual features.

In \cite{saxena2017exploring}, the network is trained to compute
relative camera pose between two images. A PBVS controller is then
employed to calculate the velocity command for a quadrotor.
Similarly in \cite{bateux2018training}, the first network
iteration is trained to find relative camera pose from a
single current image while the target image is implicitly
learned. In their attempt to generalize, the second network
iteration is modified to accept two images corresponding
to two viewpoints of the same scene and outputs the relative
camera pose between the images. The error is then passed to
a PBVS controller for servoing. This pipeline was tested on
an unseen scene and managed to achieve convergence of few centimeters. 

In \cite{yu2019siamese}, the network outputs the relative
camera pose (eye-in-hand) which can be followed as a one-shot
motion or in iterative manner. The network achieved sub-millimeter
positioning accuracy and was tested on a related but unseen VGA connector.

In \cite{tokuda2021convolutional}, the DEFINet outputs the
relative end-effector pose (eye-to-hand) and similarly
achieves sub-millimeter positioning accuracy.
More interestingly, the network was tested on four other
unseen objects at various lighting conditions and achieved
sub-millimeter positioning accuracy on most of them.
However, the position difference between the initial and
target pose is limited to few millimeters. 

Another approach, DFVS \cite{harish2020dfvs}, utilizes DNN
(based on FlowNet2 \cite{ilg2017flownet}) to output optical
flow from two images while another network estimates the depth
value for each feature.

All of the mentioned DL-based methods rely on direct comparison
of whole images instead of extracting visual features.
Thus, these approaches require the current and target images to
be essentially similar which limits generalizability. Furthermore, \cite{saxena2017exploring,bateux2018training,yu2019siamese,tokuda2021convolutional},
impose the heavy demand of estimating the full 6D pose on the DNN.
We argue that this put further limitation on generalizability.
Instead of estimating 6D pose, our network explicitly detect
features on 2D image. Arguably, detecting 2D features is easier
compared to estimating 6D pose.

\subsection{Deep Learning for Feature Extraction}

Feature extraction, more specifically for keypoints, have been
implemented in many areas such as in calibration and
localization. Traditionally, people rely on hand-crafted algorithms to
detect keypoints
\cite{lowe1999object,bay2008speeded,rublee2011orb}. The
breakthrough in DL has allowed DNN to replace the traditonal
algorithms and achieves better performance and robustness against
occlusion and variable lighting conditions. More recently, there has
been increased attention in \emph{self-supervised} network for
keypoint detection --- that is, they do not require manual human
labelling. LF-Net \cite{ono2018lf} performs training with images of
buildings but the training pipeline requires integration of
Structure-from-Motion (SfM) algorithm. SuperPoint
\cite{detone2018superpoint} pretrains the base detector using
synthetic data of simple geometries. UnsuperPoint
\cite{christiansen2019unsuperpoint} removes the pretraining
requirement and improves on the operation speed. KP2D
\cite{tang2020neural} made follow-up improvement by introducing
outlier rejection during training and better keypoint regression
method.

\subsection{Object Pose Estimation}
\label{lr:pose-est}

Object pose estimation from an RGB image is a closely related topic. One approach relies on directly estimating the 6D pose in a single shot \cite{xiang2018posecnn}. Another approach predicts 2D keypoints in the image before computing object pose through establishing 2D-3D correspondences \cite{rad2017bb8,tekin2018real,oberweger2018making}. These CNN-based methods have outperformed the traditional methods \cite{bay2008speeded,lowe1999object} in terms of accuracy and robustness to lighting and occlusion. \cite{li2018deepim,peng2019pvnet} are some of the best performing methods which score highly on the LINEMOD dataset \cite{hinterstoisser2012model}. One of the evaluation metrics that they use is the 2D projection metric. With this, they measure the difference between projected object's vertices in the 2D image between ground-truth and the estimated object pose. The estimated pose is considered correct if the average pixel difference is below 5 pixels (for 480x640 image). Another well-known metric is the \textit{5\degree}, \textit{5cm} metric where the estimated pose is considered correct if the rotation is under 5\degree and translation under 5 cm respectively.

\section{Background on IBVS}
\label{part:vs-intro}

Here we provide a short introduction to IBVS that is relevant to our control application. For more explanation, we refer the reader to \cite{chaumette2006visual}.

The goal of IBVS is to minimize the error between features obtained in the image space.
\vspace{-0.1cm}
\begin{align}
\boldsymbol{\mathrm{e}}(t) = \boldsymbol{\mathrm{s}}(\boldsymbol{\mathrm{p}}(t),\boldsymbol{\xi}) - \boldsymbol{\mathrm{s}}^*(\boldsymbol{\mathrm{p}}^{*},\boldsymbol{\xi}) \label{eq1}
\end{align}

In the equation above, $\boldsymbol{\mathrm{p}}=(u_1,v_1,...,u_k,v_k) \in \mathbb{R}^{2k}$ contains the set of 2-D pixel coordinates of $k$ features in the current image $\boldsymbol{\mathrm{I}}$. The pixel values are converted to the image plane cartesian coordinate $\boldsymbol{\mathrm{s}} = \boldsymbol{\mathrm{s}}(\boldsymbol{\mathrm{p}}(t),\boldsymbol{\xi}) = (x_1,y_1,...,x_k,y_k) \in \mathbb{R}^{2k}$ using the set of camera intrinsic parameters $\boldsymbol{\xi}$. Similarly, $\boldsymbol{\mathrm{p}}^{*}$ and $\boldsymbol{\mathrm{s}}^{*}$ are measurements obtained from the target image $\boldsymbol{\mathrm{I}}^*$. 

The spatial velocity expressed in camera frame, or camera twist, is defined as $\boldsymbol{\mathrm{v}}_c = (v_x,v_y,v_z,w_x,w_y,w_z) \in \mathbb{R}^{6}$ and is related to $\boldsymbol{\mathrm{s}}$ by the equation:
\vspace{-0.1cm}
\begin{align}
\boldsymbol{\dot{\mathrm{s}}} = \boldsymbol{\mathrm{L}}(\boldsymbol{\mathrm{s}},\boldsymbol{\mathrm{Z}})\boldsymbol{\mathrm{v}}_c \label{eq2}
\end{align}

\noindent where $\boldsymbol{\mathrm{L}}(\boldsymbol{\mathrm{s}},\boldsymbol{\mathrm{Z}}) \in \mathbb{R}^{2k\times6}$ is the interaction matrix, and $\boldsymbol{\mathrm{Z}} \in \mathbb{R}^{k}$ is the feature depth vector.

From \eqref{eq1} and \eqref{eq2}, we obtain:
\vspace{-0.1cm}
\begin{align}
\boldsymbol{\dot{\mathrm{e}}} = \boldsymbol{\mathrm{L}}(\boldsymbol{\mathrm{s}},\boldsymbol{\mathrm{Z}})\boldsymbol{\mathrm{v}}_c \label{eq3}
\end{align}

To achieve an exponential decoupled decrease of error, a classical IBVS control law  \cite{chaumette2006visual} is adopted as follows:
\vspace{-0.1cm}
\begin{align}
  \boldsymbol{\mathrm{v}}_c = -\lambda \boldsymbol{\widehat{\mathrm{L}}}^+(\boldsymbol{\mathrm{s}},\boldsymbol{\mathrm{Z}}) \boldsymbol{\mathrm{e}} \label{eq4}
\end{align}

\noindent where $\boldsymbol{\widehat{\mathrm{L}}}$ is an approximation of $\boldsymbol{\mathrm{L}}$. There are several ways to construct $\boldsymbol{\widehat{\mathrm{L}}}$ \cite{chaumette2006visual}. In our implementation, we use $\boldsymbol{\widehat{\mathrm{L}}}(\boldsymbol{\mathrm{s}},\boldsymbol{\mathrm{Z}}) = \boldsymbol{\mathrm{L}}_{*}(\boldsymbol{\mathrm{s}}^*,\boldsymbol{\mathrm{Z}}^*)$ which is the value of $\boldsymbol{\mathrm{L}}$ at the target pose as captured in $\boldsymbol{\mathrm{I}}^*$.

\section{Deep Feature-Based Visual Servo (DFBVS)}
\label{part:dfbvs}

\subsection{Render Engine}
\label{part:dfbvs-render}

In robotic \textit{grasping} or \textit{inspection} operation, typically the target end-effector or camera pose relative to the target object is specified from a motion planner. However, converting the target pose to target image for visual servoing is riddled in paradox: if one can operate the robot accurately to obtain the target image $\boldsymbol{\mathrm{I}}^*$, there is no longer a need for visual servoing. When collaborative robot is in the picture, hand-guiding for $\boldsymbol{\mathrm{I}}^*$ becomes a feasible option albeit highly challenging when high accuracy is required.

The implementation of an image render engine allows us to circumvent said paradox quickly. We apply Eevee renderer in Blender 3D software to directly generate $\boldsymbol{\mathrm{I}}^*$ from the target pose. The render engine also simplifies the requirement for depth values of features in constructing $\boldsymbol{\widehat{\mathrm{L}}}$. By choosing $\boldsymbol{\widehat{\mathrm{L}}} = \boldsymbol{\mathrm{L}}_{*}$, the target feature depth vector $\boldsymbol{\mathrm{Z}}^*$ can be easily obtained from the render engine.

\subsection{Feature Extraction}
\label{part:dfbvs-detect}

\begin{table}[t]
\centering
\vspace{1mm}
\caption{Experiment result for robotic grasping on three objects}
\label{tab:sogt-1}
\begin{tabular}{p{0.15\columnwidth}P{0.26\columnwidth}P{0.25\columnwidth}P{0.05\columnwidth}P{0.05\columnwidth}}
 \hline
 experiment & $|\Delta x$,$\Delta y$,$\Delta z|$ [mm] & $|\Delta rx$,$\Delta ry$,$\Delta rz|$ [deg] & AVG1 [px] & AVG2 [px]\\
 \hline
 O$_1$G$_1$T$_1$ & 3.25, 3.49, 67.72 & 10.42, 6.70, 0.13 & 0.71 & 0.78\\
 O$_1$G$_1$T$_2$ & 2.88, 0.13, 86.04 & 15.01, 3.78, 0.17 & 1.14 & 1.19\\
 O$_1$G$_1$T$_3$ & 21.22, 11.45, 64.94 & 7.61, 9.39, 7.87  & 1.00 & 1.28\\
 \hline
 O$_1$G$_2$T$_1$ & 13.86, 16.90, 49.17 & 4.27, 6.28, 4.74 & 0.89 & 1.07\\
 O$_1$G$_2$T$_2$ & 3.12, 12.25, 51.63 & 6.78, 3.66, 1.05 & 0.74 & 1.12\\
 O$_1$G$_2$T$_3$ & 3.81, 14.94, 57.53 & 6.86, 4.61, 1.31 & 1.04 & 1.39\\
 \hline
 O$_2$G$_1$T$_1$ & 1.00, 12.36, 59.18 & 7.67, 2.99, 0.74 & 0.85 & 1.13\\
 O$_2$G$_1$T$_2$ & 1.53, 24.47, 44.40 & 4.21, 1.57, 1.21 & 0.74 & 0.56\\
 O$_2$G$_1$T$_3$ & 1.15, 10.33, 24.12 & 2.85, 0.40, 0.86 & 0.73 & 0.49\\
 \hline
 O$_2$G$_2$T$_1$ & 0.18, 11.08, 41.28 & 5.07, 1.98, 0.42 & 0.62 & 0.46\\
 O$_2$G$_2$T$_2$ & 1.49, 27.51, 43.66 & 3.77, 1.90, 0.96 & 0.65 & 0.59\\
 O$_2$G$_2$T$_3$ & 5.90, 11.79, 51.78 & 5.97, 3.70, 1.62 & 0.81 & 0.59\\
 \hline
 O$_3$G$_1$T$_1$ & 27.85, 4.72, 51.49 & 10.88, 5.29, 1.76 & 1.10 & 0.94\\
 O$_3$G$_1$T$_2$ & 0.58, 12.18, 47.08 & 8.56, 0.12, 2.97 & 0.68 & 0.74\\
 O$_3$G$_1$T$_3$ & 15.51, 9.30, 56.42 & 9.48, 4.73, 5.41 & 0.63 & 0.94\\
 \hline
 O$_3$G$_2$T$_1$ & 2.64, 20.5, 34.79 & 4.93, 0.67, 3.34 & 0.86 & 0.89\\
 O$_3$G$_2$T$_2$ & 8.61, 23.25, 49.68 & 5.15, 7.13, 1.84 & 1.03 & 1.16\\
 O$_3$G$_2$T$_3$ & 1.30, 20.28, 49.38 & 7.58, 1.83, 2.44 & 0.83 & 0.74\\
 \hline
\end{tabular}
\end{table}

The KeyPointNet network \cite{tang2020neural}, $K : \boldsymbol{\mathrm{I}} \rightarrow \{\boldsymbol{\mathrm{p}},\boldsymbol{\mathrm{f}},\boldsymbol{\mathrm{c}}\}$, takes an input image and outputs a set of $n$ keypoints $\boldsymbol{\mathrm{p}}$ (similar in \eqref{eq1}), descriptors $\boldsymbol{\mathrm{f}} \in \mathbb{R}^{n\times256}$, and normalized confidence score $\boldsymbol{\mathrm{c}} \in \mathbb{R}^{n}$.

The network is trained in a self-supervised manner with weighted combination of loss functions. The authors claimed that this allowed them to enforce keypoint detection at different views, minimize distance between descriptors of the same feature seen at different views, maximize distance between different features and enforce consistent score from different views. We refer to the original paper for more details.
 

\subsection{Feature Matching and Outlier Rejection}
\label{part:dfbvs-match}

At each control cycle $t$, the features $\boldsymbol{\mathrm{s}}(t)$ and $\boldsymbol{\mathrm{s}}^*$ are matched based on the descriptor values to form $n$ correspondences $\boldsymbol{\mathrm{C}}_t$ via \textit{nearest-neighbour} function $N$:
\vspace{-0.1cm}
\begin{align}
N:\{\boldsymbol{\mathrm{s}}(t),\boldsymbol{\mathrm{f}}(t),\boldsymbol{\mathrm{s}}^*,\boldsymbol{\mathrm{f}}^*\} \rightarrow \boldsymbol{\mathrm{C}}_t \label{eqN}\\
\boldsymbol{\mathrm{C}}_t = [\boldsymbol{\mathrm{m}}, \boldsymbol{\mathrm{m}}^*] \in \mathbb{R}^{n\times4}
\end{align}

where $\boldsymbol{\mathrm{m}}$ and $\boldsymbol{\mathrm{m}}^*$ are simply $\boldsymbol{\mathrm{s}}(t)$ and $\boldsymbol{\mathrm{s}}^*$ respectively which are rearranged to form feature pairs which are close in the descriptor space.

\begin{figure}
  \centering
  \vspace{1mm}
  \includegraphics[width=\columnwidth]{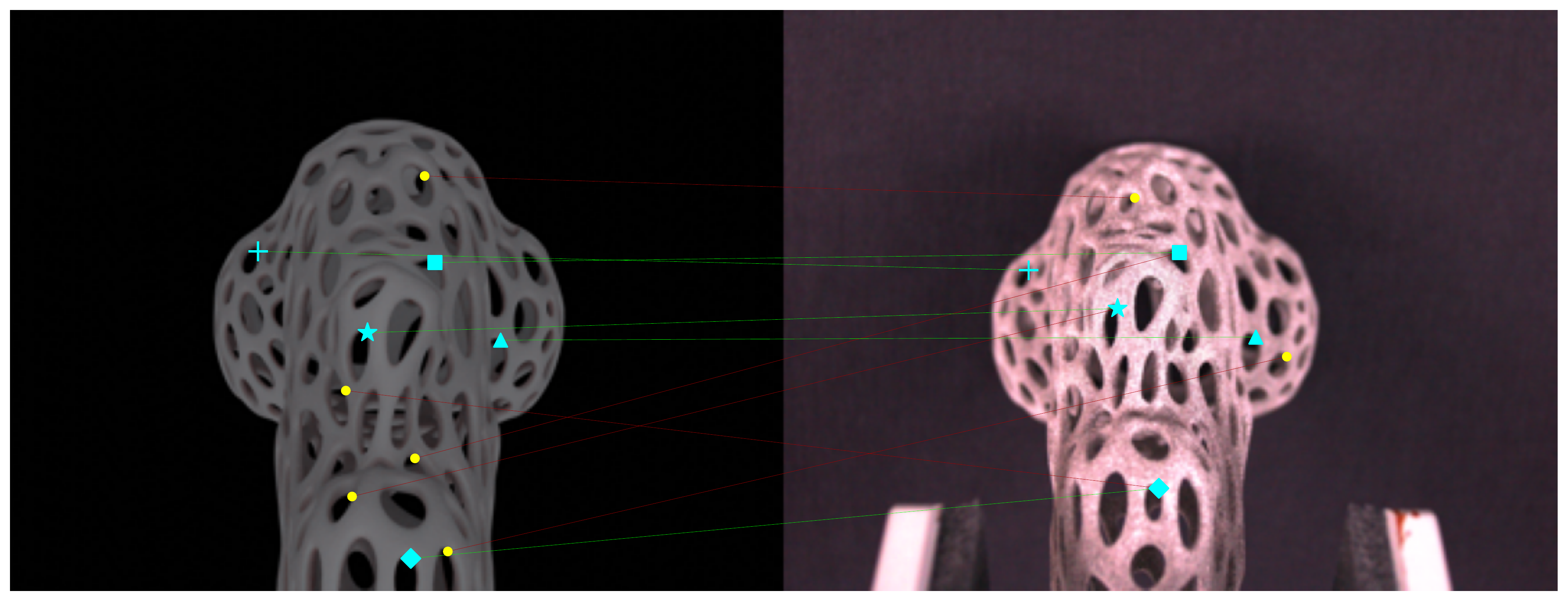}
  \caption{Blue points indicate 3D points which are detected in both target and current images. Each pair of unique 3D point is distinguished by similar marker shape in both images. Points in yellow indicate 3D points which are not detected in the other image i.e. has no true correspondence. Performing correspondence matching will result in true correspondences shown in green and false correspondences shown in red.}
  \label{fig:sogt_ablation_ransac}
\end{figure}

Since $\boldsymbol{\mathrm{s}}(t)$ and $\boldsymbol{\mathrm{s}}^*$ are obtained from different images, they do not necessarily have a one-to-one match (e.g. a feature in $\boldsymbol{\mathrm{I}}(t)$ might not be visible in $\boldsymbol{\mathrm{I}}^*$). Therefore, the correspondence matching in (\ref{eqN}) is bound to produce some false correspondences. Refer to Figure \ref{fig:sogt_ablation_ransac} for an illustration.

In order to mitigate this, we implement RANSAC \cite{fischler1981random} ($R : \boldsymbol{\mathrm{C}}_t \rightarrow \boldsymbol{\mathrm{R}}_t$, $\boldsymbol{\mathrm{R}}_t \subseteq \boldsymbol{\mathrm{C}}_t$ ) on the correspondences to estimate inlier subset $\boldsymbol{\mathrm{R}}_t$ that will be used as input to the control law.
\vspace{-0.1cm}
\begin{align}
\boldsymbol{\mathrm{R}}_t = [\boldsymbol{\mathrm{r}}, \boldsymbol{\mathrm{r}}^*] \in \mathbb{R}^{l\times4}
\end{align}

where $\boldsymbol{\mathrm{r}} \subseteq \boldsymbol{\mathrm{m}}$, $\boldsymbol{\mathrm{r}}^* \subseteq \boldsymbol{\mathrm{m}}^*$, and $l \leq n$.

\subsection{Establishing Tracking Behavior}
\label{part:vs-pseudo-tracking}

\begin{figure}
  \centering
  \vspace{1mm}
  \includegraphics[width=\columnwidth]{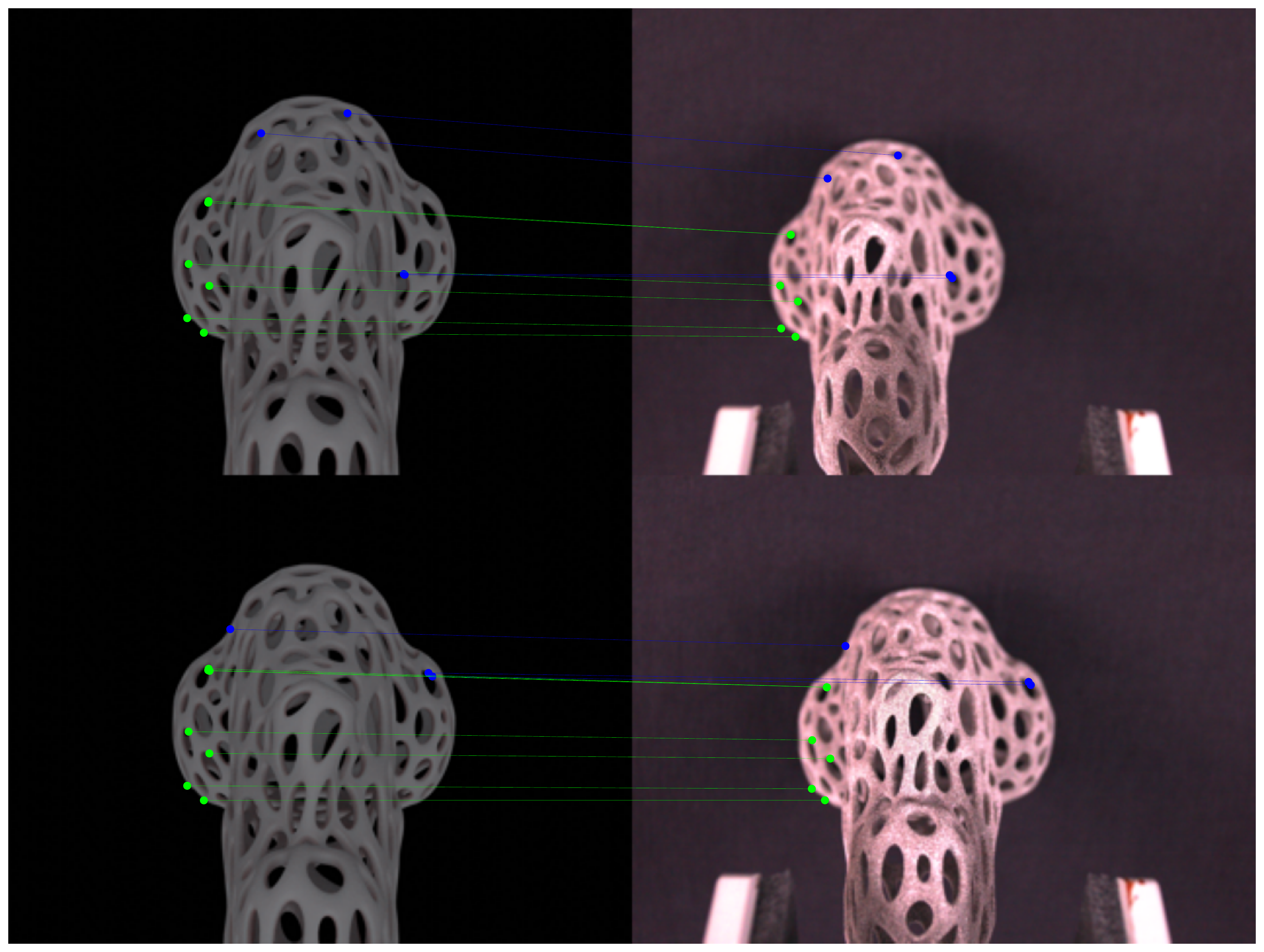}
  \caption{(First row) Correspondences at cycle $t-1$. (Second row) Correspondences at cycle $t$. (Correspondence-switch) Only the correspondences in green consistenly appear in both cycles. The correspondences in blue have undergone \emph{correspondence-switching}.}
  \label{fig:soft_ablation_switchcorr}
\end{figure}

At each cycle, the set of 3D points that forms the correspondences might not be identical with the set in the previous or following cycle. When this happens, we say that the system displays \emph{correspondence-switching} behavior. Refer to Figure \ref{fig:soft_ablation_switchcorr} for an illustration.

This becomes a problem when the camera pose is close to the target pose as the \emph{correspondence-switching} causes oscillations that complicates convergence. In order to mitigate this, we enforce \emph{tracking} behavior when the camera image is close to $\boldsymbol{\mathrm{I}}^*$ (we estimate this by taking the average of $\boldsymbol{\mathrm{e}}(t)$). We notate cycles with \emph{tracking} behavior as $t'$.

At $t'$, the correspondences are always a subset of the previous correspondences ($\boldsymbol{\mathrm{R}}_{t'} \subseteq \boldsymbol{\mathrm{R}}_{t'-1}$). This is obtained by matching features from current image ($\boldsymbol{\mathrm{s}}(t')$) with target features which are present in previous cycle ($\boldsymbol{\mathrm{r}}^*$ at $t'-1$) to obtain the correspondence set $\boldsymbol{\mathrm{C}}_{t'}$ and the inlier set $\boldsymbol{\mathrm{R}}_{t'}$ respectively.


\section{Experiment}
\label{part:exp}

\begin{figure}
  \centering
  \vspace{2mm}
  \includegraphics[width=\columnwidth]{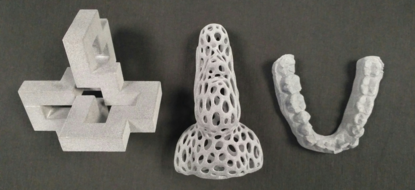}
  \caption{Three objects O$_1$, O$_2$, and O$_3$ which were 3D printed using HP MJF5200.}
  \label{fig:objects}
\end{figure}

\begin{figure}[h!]
  \centering
  \begin{subfigure}{\columnwidth}
    \includegraphics[height=3.2cm]{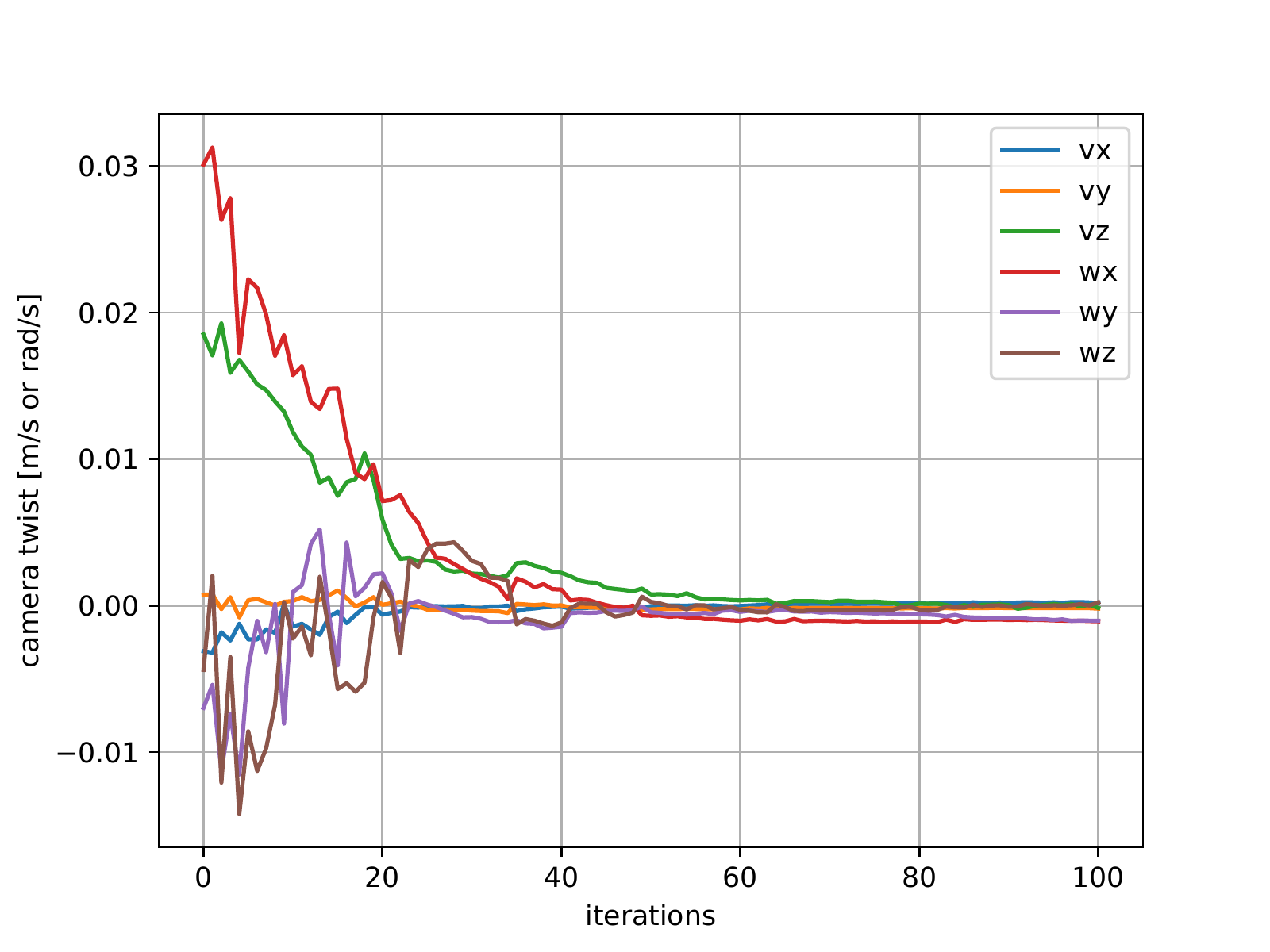}
    \includegraphics[height=2.9cm]{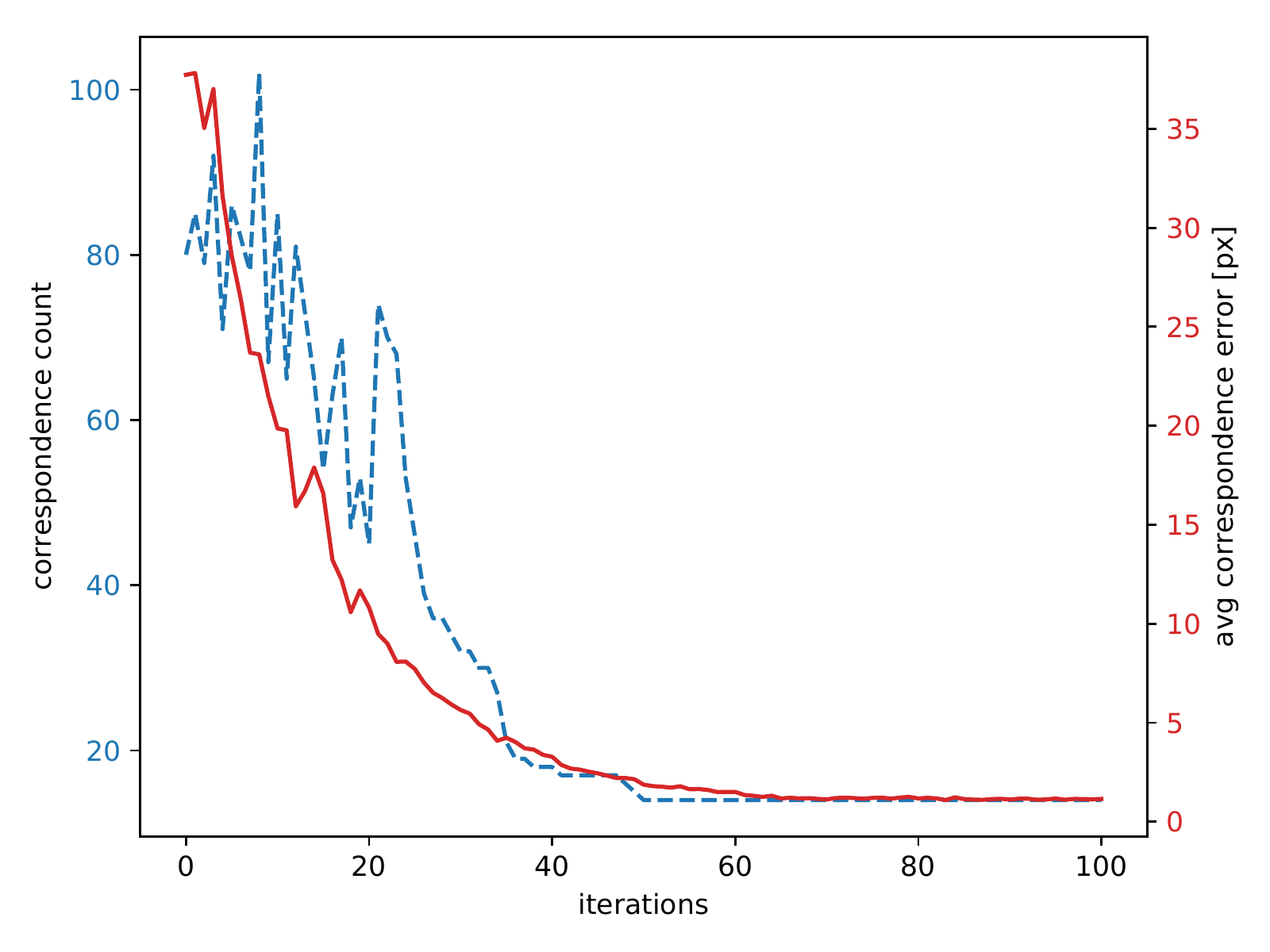}
    \caption{}
    \label{fig:o1-profiles}
  \end{subfigure}
  \begin{subfigure}{\columnwidth}
    \includegraphics[height=3.2cm]{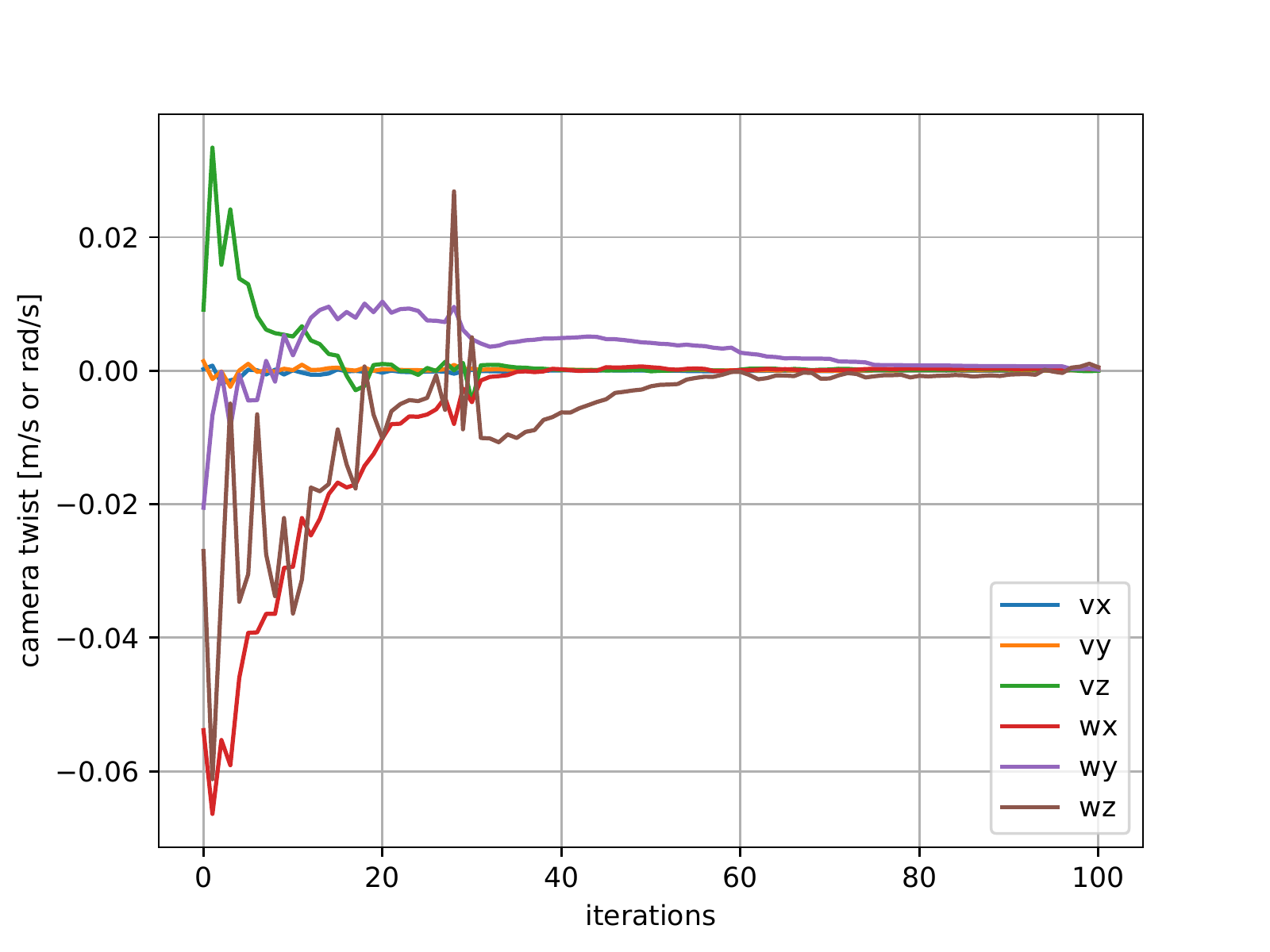}
    \includegraphics[height=2.9cm]{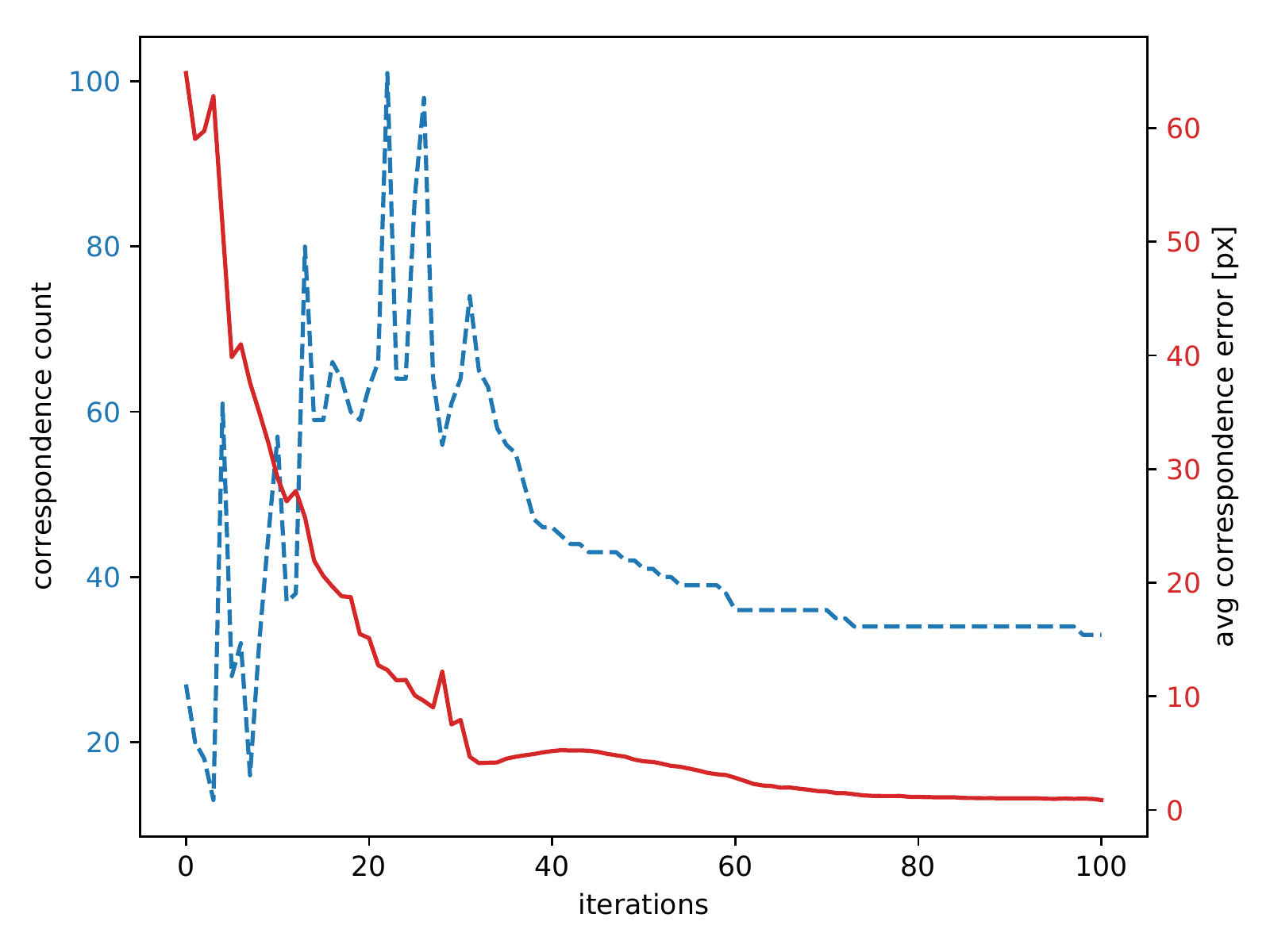}
    \caption{}
    \label{fig:o2-profiles}
  \end{subfigure}
  \begin{subfigure}{\columnwidth}
    \includegraphics[height=3.2cm]{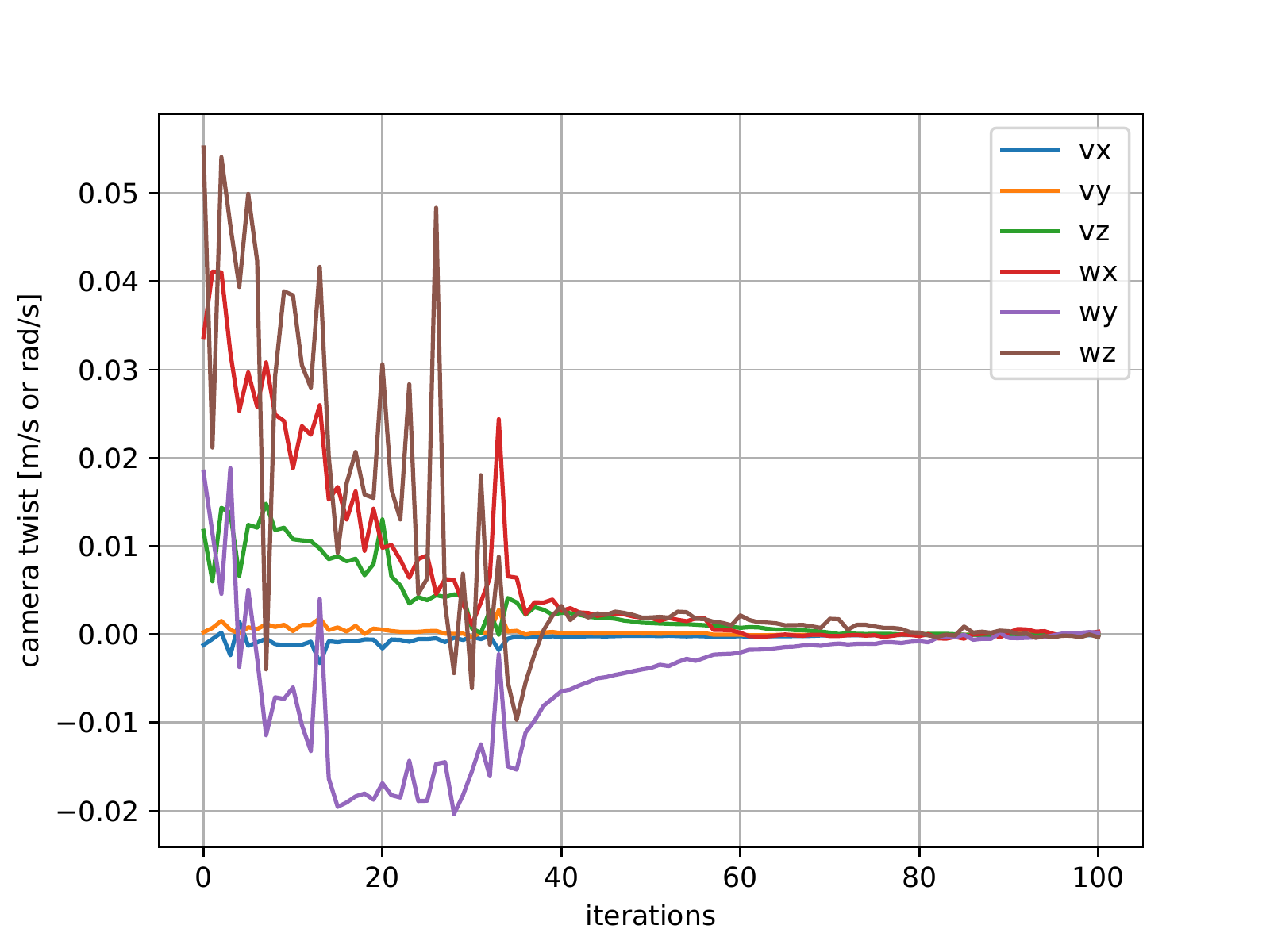}
    \includegraphics[height=2.9cm]{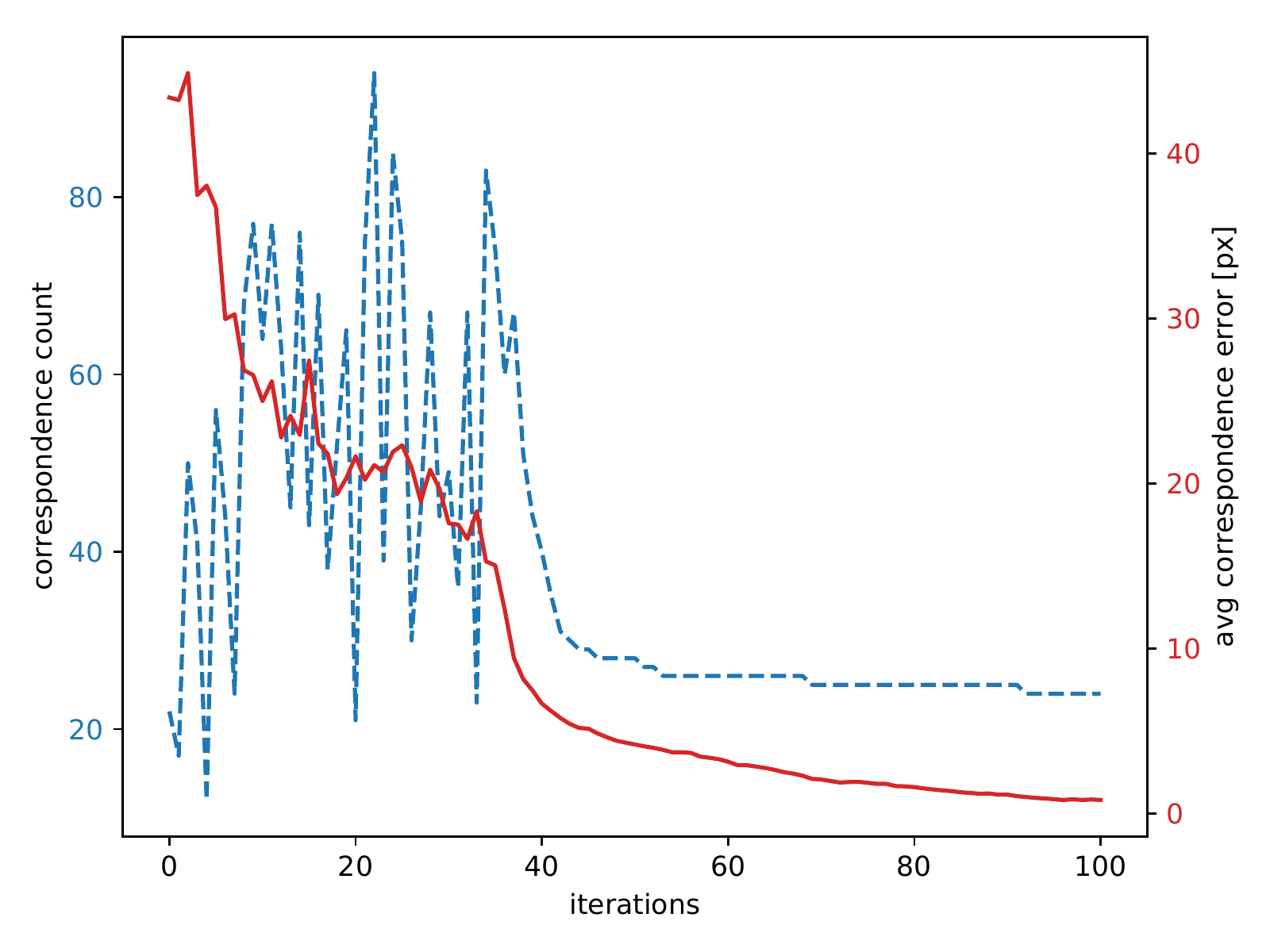}
    \caption{}
    \label{fig:o3-profiles}
  \end{subfigure}
  \caption{Each row shows profiles from visual servoing on different unseen objects: (a) O$_1$ (b) O$_2$ (c) O$_3$. Figures on the left are camera twist profiles. Figures on the right show average correspondence error in red and correspondence count in blue.}
  \label{fig:profiles}
\end{figure}

\begin{figure*}[h!]
  \centering
  \vspace{2mm}
  \includegraphics[width=\textwidth]{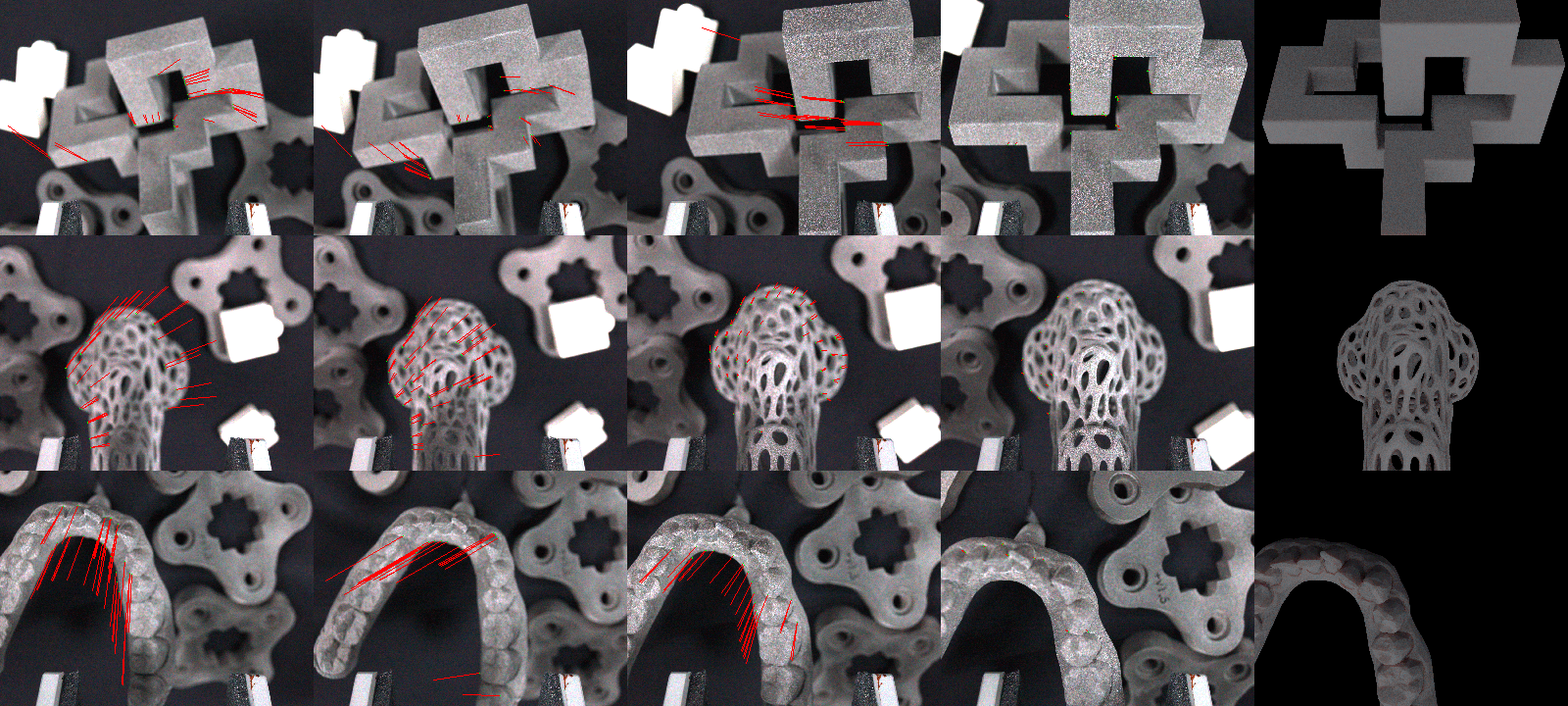}
  \caption{Sequence of camera images from visual servoing experiments on cluttered environment with DFBVS. The feature errors, which are used as input to the IBVS control law, are displayed as red lines. The final camera images are shown on the fourth column, showing alignment with the rendered target images on the fifth column. Row 1, 2, and 3 are of object O$_1$, O$_2$, and O$_3$ respectively. Video of the experiments is available at \url{https://youtu.be/vFAdebgog7k}.}
  \label{fig:collage-cos}
\end{figure*}

\begin{figure}[h!]
  \centering
  \includegraphics[width=\columnwidth]{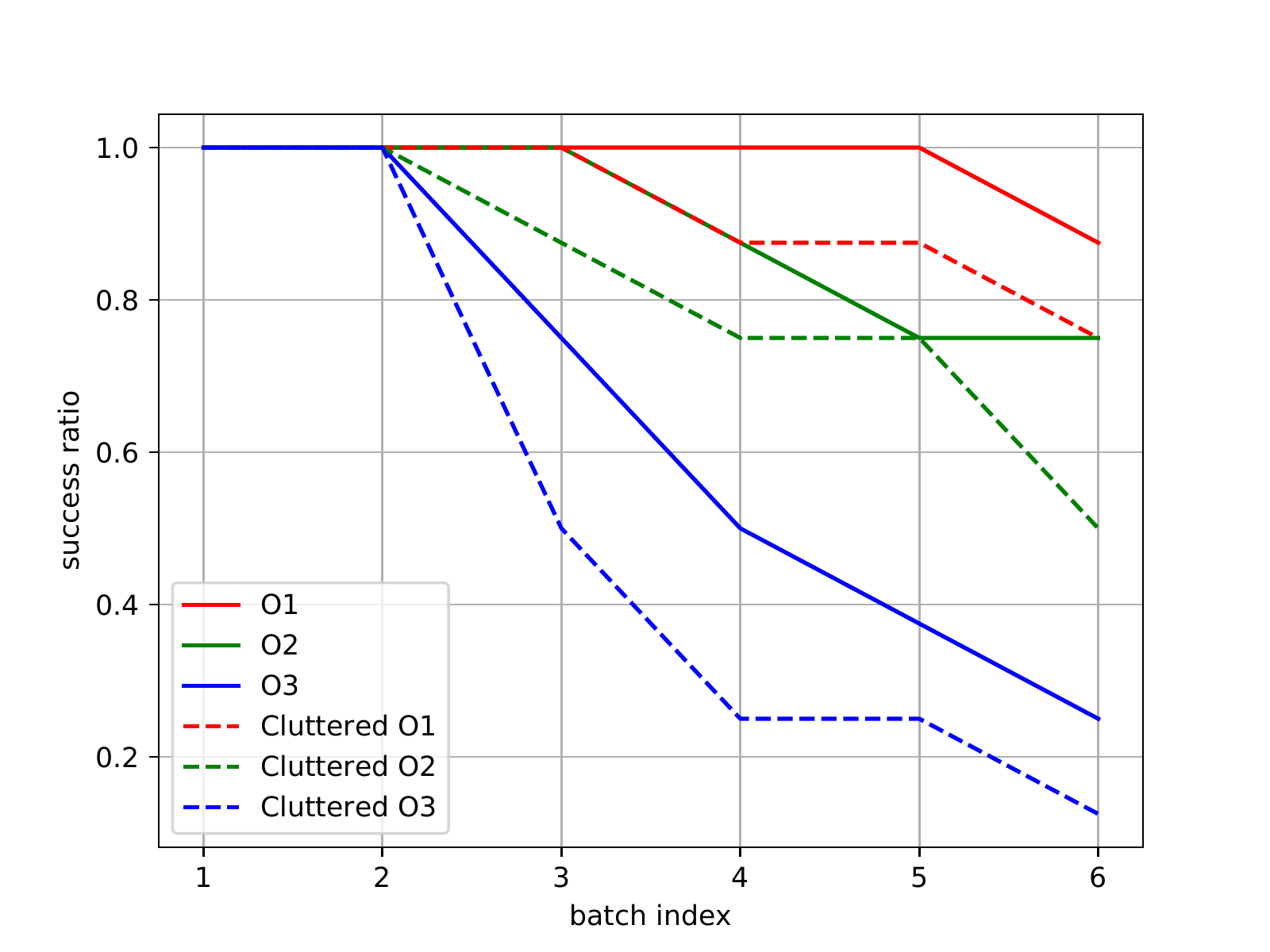}
  \caption{Plot of grasping success ratio at each of the six batches where the distance the camera has to follow increases linearly from within 1 to 6 cm. Results for object O$_1$,O$_2$,O$_3$ are in red, green, and blue respectively. Experiments with similar backgrounds are in bold lines. Experiments with more challenging dissimilar backgrounds are in dashed lines. Our method achieved 100\% convergence on all cases for batch 1 and 2 (within 2 cm distance).}
  \label{fig:grasp-ratio}
\end{figure}

We run several experiments to assess the performance of DFBVS. The setup consists of a FLIR BFS-U3-120S4C camera with 12 mm lens installed on the end-effector of a Universal Robot UR5e manipulator.

Image from the camera is downsized to 320x240 before being forwarded through the network. We chose the top 500 keypoints from the network output ranked by the confidence score $\boldsymbol{\mathrm{c}}$. To showcase the generalizability of the pipeline, we apply the same network weight (v4), which was trained on COCO dataset \cite{lin2014microsoft} by the original paper's authors \cite{tang2020neural}, without any fine-tuning. Experiments are done on unseen 3D printed objects (O$_1$,O$_2$,O$_3$) as seen in Figure \ref{fig:objects}. The control section is implemented with ViSP library \cite{Marchand05b}.

\subsection{Accuracy Assessment}
\label{part:exp-acc}

In this section, we perform experiments to quantify the accuracy of the visual servoing pipeline. As the target image was obtained through a renderer, there is no ground-truth data available on the target robot configuration. Here, we use accuracy in the image space as the accuracy metric as inspired from pose estimation field \cite{li2018deepim,peng2019pvnet}.

Two goal poses (G$_1$,G$_2$) expressed in the object frame are identified
for each object. We visit each goal pose from three initial poses
(T$_1$,T$_2$,T$_3$). In total there are 18 (3x2x3) unique servoing
arrangements for the three objects. The absolute positional distance
for servoing averaged 55.1 $\pm$ 12.2 mm and 8.7 $\pm$ 3.4 deg for orientation.

We list the experiment results in Table \ref{tab:sogt-1}. For each experiment, we present two measurements of the average final correspondences error in pixel units. The first one (AVG1) is the average error calculated from all the correspondences in the final set $\boldsymbol{\mathrm{R}}_{final}$. For the second measurement (AVG2) we carefully select three correspondences from $\boldsymbol{\mathrm{R}}_{final}$ and report the average error. The three correspondences are selected based on the following criteria:
\begin{itemize}
\item every correspondence pair is visually examined to point to the same 3D point
\item the three correspondences to be adequately distanced from one another
\end{itemize}

Looking at AVG2, the average error is 0.89 pixels which is lower than the threshold commonly used in pose estimation field (5 pixels) to categorize successful estimation (as mentioned in Section \ref{lr:pose-est}).

In Figure \ref{fig:profiles} we show example visual servoing profiles for each object to show the system's ability to converge. The convergence can be attributed to the \emph{tracking} strategy (indicated by later iterations with non-increasing correspondence count).

\subsection{Integration in a Grasping Pipeline}
\label{part:exp-grasp-ratio}

For the second experiment, we assess our pipeline's effectiveness for robotic grasping. The experiment mimics a typical grasping operation in a 3D print factory. Freshly printed objects would have to be grasped from a particular pose in the environment, which is likely cluttered.

The experiments are divided into six batches where the distance that the camera has to travel increases linearly from within 1 to 6 cm, beyond which the camera image gets too blurry. The rotations along X,Y,Z axes are varied within the range [8\degree,6\degree,3\degree], [6\degree,6\degree,4\degree], and [8\degree,10\degree,8\degree] for O$_1$, O$_2$, O$_3$ respectively. Eight experiments are performed at each stage. A run is considered successful if the average correspondences error falls below 2 pixels. The background in the actual setup is kept to a simple dark background and no elaborate \textit{tuning} was done to adjust the rendered lighting. The sharpness of the camera image is also not enforced throughout the runs. Note that there are still notable discrepancies between the two input images such as appearance of the gripper fingertips in the camera image.

In addition to the above, we run another set of experiments with similar setup except now we clutter the real environment with other objects. Meanwhile, the background of the rendered images remain empty.

Both results are presented in Figure \ref{fig:grasp-ratio}. Results for first experiment set are shown in bold lines while results for the second set are shown in dashed lines. Note that in all cases, the method can achieve perfect convergence when the distance between initial and final pose is within 2 cm, which can be achieved through rough positioning. Example images from the second set of experiment are displayed in Figure \ref{fig:collage-cos}.

\section{Conclusion}
\label{part:conclusion}

Most recent approaches in visual servoing remove reliance on visual features and instead compare directly the target and current camera images. However, we argue that this comes at a cost of generalizability.
Our proposed approach perform VS based on visual features as in
classical VS approaches but, contrary to the latter, we leverage
recent breakthroughs in Deep Learning to automatically extract and
match the visual features. This allows our method to be generalizable
to unknown, cluttered, scenes. Such generalizability furthermore allows
our method to integrate a render engine to generate synthetic target
image. This enables a simple and automatic way to acquire the target
image, while providing accurate depth values for the visual features.
For future works, we will look into applying the pipeline for assembly tasks and improving the control performance.

\section*{Acknowledgment}
This study is supported under the RIE2020 Industry Alignment Fund – Industry Collaboration Projects (IAF-ICP) Funding Initiative, as well as cash and in-kind contribution from the industry partner, HP Inc., through the HP-NTU Digital Manufacturing Corporate Lab.

\bibliographystyle{ieeetr}
\bibliography{dfbvs}
\end{document}